\documentclass[letterpaper]{article} 
\usepackage[preprint]{aaai2027}
\nocopyright 
\usepackage[hyphens]{url}  
\usepackage{graphicx} 
\usepackage{natbib}  
\usepackage{caption} 
\usepackage{booktabs}

\title{ToolGate: An Executable Acceptance Pipeline for Tool-Dependent Scientific Benchmark Construction}

\author{
Ke Zhang, Yankang Liu \\
Roya Zandi, Maziar Raissi
}

\affiliations{
University of California, Riverside\\
Riverside, CA, USA\\
kzhan153@ucr.edu, yliu937@ucr.edu, royaz@ucr.edu, maziar.raissi1@ucr.edu
}

\begin{document}

\maketitle

\begin{abstract}
Scientific benchmarks are commonly built by domain experts who write tasks and
cross-check one another's work, or who adapt existing material from textbooks,
published papers, and online resources. These routes can produce strong
evaluations, but they require substantial per-item labor. Language models can
reduce this repeated work by
proposing candidates quickly. The remaining problem is acceptance. We target
scientific questions whose answers require computations with specialist
software rather than unaided reasoning alone. A candidate is invalid if its script fails or returns a different answer,
or trivial if a model answers it without the software. We present ToolGate, which treats every generated
item as a proposal and keeps it only if three gates pass. First, an executable
solution script must reproduce the proposed answer when run with the scientific
software. Second, randomized no-tool screening rejects candidates that models
can already solve from the prompt alone. Third, a tool-using agent must solve each survivor within a fixed time
limit. We
instantiate ToolGate in FEniCSx with 500 generation attempts. The
local-verification gate retains 478 candidates. For final reporting, we
rescreen this pool after generation: two randomized no-tool screens exclude
222 from the reported pool, and direct GPT-5.5 API calls at medium reasoning (the API default)
exclude another 121. Of
the remaining 135, a GPT-5.5 Codex CLI
agent with access to FEniCSx solves 130; exact
deduplication leaves 128 unique protocol survivors. ToolGate turns repeated
answer checking and difficulty screening into an auditable process while
leaving domain design and final review to experts.
\end{abstract}

\section{Introduction}

Scientific benchmarks are commonly built by domain experts who write tasks and
cross-check one another's work, or who adapt existing material. GPQA asks
domain experts to write questions and other experts to solve and validate
them~\citep{rein2024gpqa}. Scientific-agent benchmarks including
ScienceAgentBench, SciCode, and CORE-Bench adapt tasks, data, and code from
published research, followed by expert curation and
review~\citep{chen2024scienceagentbench,tian2024scicode,siegel2024corebench}.
These approaches can produce strong evaluations, but they repeat expensive
work for every item: writing or extracting a task, solving it, and
independently checking the result.

Language models can reduce this labor by quickly proposing a question, an
answer, and an attempted solution. These outputs still need to be checked. The
attempted solution may fail or return a different answer. The question may
also be too easy: a model may
answer it from wording, prior knowledge, or reasoning alone. Such a question is
not useful for evaluating whether a model can use scientific software. Prior
work addresses related problems in other settings. APIGen and AutoCodeBench
execute and filter generated data to check correctness, while AutoBencher
measures and optimizes question
difficulty~\citep{liu2024apigen,chou2025autocodebench,li2024autobencher}.
Generating candidates quickly therefore does not solve the main problem: we
still need to check each one and decide whether to keep it.

We present ToolGate, a pipeline that checks each generated candidate before
accepting it. In our study, each candidate contains a question, four answer
options, a proposed answer, and a solution script that calls scientific
software. The multiple-choice format lets every stage compare its result with
the same proposed answer automatically. ToolGate uses three gates
(Figure~\ref{fig:pipeline}). First, it runs the script and requires its result
to agree with the proposed answer. Second, it asks models to answer using only
the question and options, without access to the scientific software, and
rejects candidates those models can solve. Third, it gives the remaining
candidates to an agent that can write and run code with the software, and keeps
those it solves. A candidate passes only if it meets all three conditions.

ToolGate can be used when scientific software can be called from code and
produces outputs that can be checked automatically. We test it with the FEniCSx
finite-element software stack~\citep{baratta2023dolfinx}.
FEniCSx is a widely used open-source platform for solving partial differential
equations with the finite-element method. A user describes meshes, function
spaces, and variational forms in Python, and FEniCSx assembles and solves the
resulting numerical systems. Our questions ask for exact outputs from multistep
FEniCSx computations, so the intended solution is to write and run code.

Of 500 generated candidates, 478 have scripts that reproduce their proposed
answers. For final reporting, we rescreen these candidates after generation.
Two randomized no-tool screens exclude 222 from the reported pool, leaving
256. Direct GPT-5.5 API calls at medium reasoning (the API default), without access to FEniCSx,
exclude another 121. Of
the remaining 135, a GPT-5.5 Codex CLI
agent with access to FEniCSx solves 130 under the reported test settings. These
results show that
asking an LLM to generate a tool-dependent question is not enough. We must test
whether the specified models fail without the software and whether a
tool-enabled agent succeeds with it.

These measurements are reliable only if the screening protocol is sound. We
show that a fixed option order lets a model's letter preferences masquerade
as question difficulty, and that unconstrained generation repeatedly produces
a few templates that often pass; the reported screens therefore randomize option order
and grade by answer value, and exact duplicate questions are removed before
release.

Our contributions are:
\begin{itemize}
\item ToolGate, which accepts a generated candidate only if its script
  reproduces the proposed answer, none of the specified no-tool screens
  answers it by majority, and the specified tool-enabled agent
  succeeds;
\item a 500-attempt FEniCSx study in which 478 candidates pass local
  verification, followed by post-generation rescreening in which two
  randomized no-tool screens exclude 222 from the reported pool, a GPT-5.5
  no-tool API screen at medium reasoning excludes another 121, and a GPT-5.5 Codex CLI agent
  with access to FEniCSx solves 130 of the remaining 135, leaving 128 unique
  protocol survivors after exact deduplication; and
\item an analysis showing how the screening model and answer presentation
  change which candidates survive, together with randomized value grading and
  exact deduplication as practical safeguards.
\end{itemize}

\section{ToolGate}
\label{sec:method}

\begin{figure*}[t]
\centering
\includegraphics[width=\textwidth]{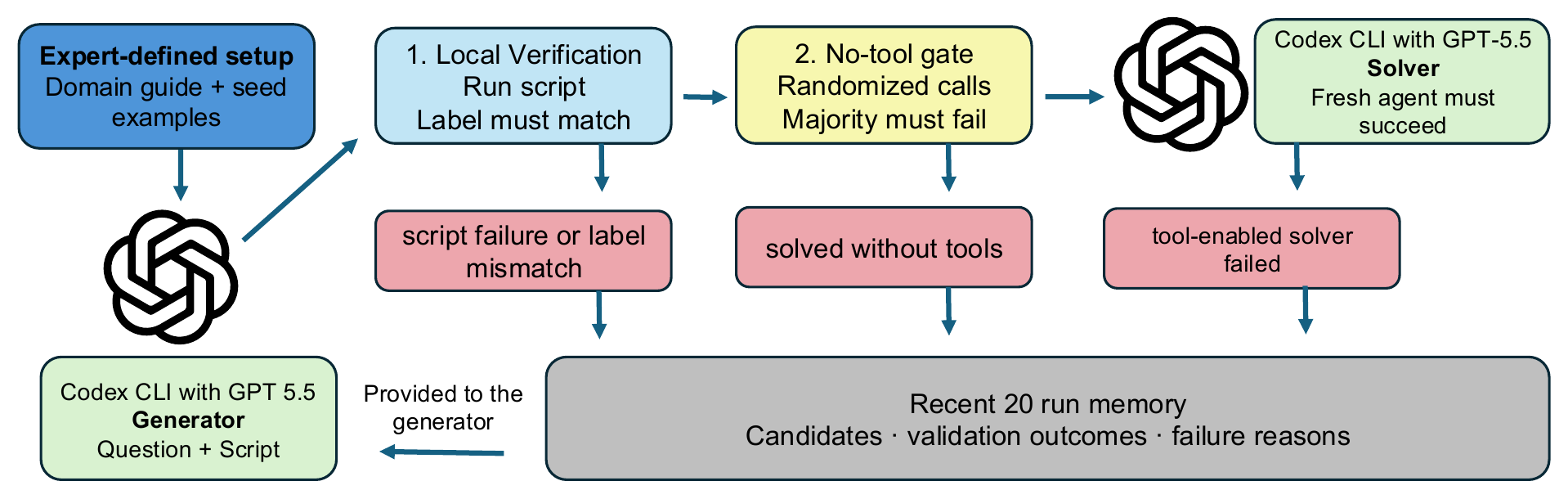}
\caption{The ToolGate loop. A generator proposes a complete
candidate---question, options, proposed answer, and solution script---and three
gates decide whether to keep it. Each gate leaves a machine-readable record,
and every outcome is appended to the run database.}
\label{fig:pipeline}
\end{figure*}

ToolGate treats every model output as a candidate rather than an accepted
benchmark item. Each candidate contains a question, four answer options, the
generator's proposed answer, and a solution script that uses the target
scientific software. A generator proposes the candidate, three gates test it,
and the outcome is saved for the next generation round
(Figure~\ref{fig:pipeline}).

\subsection{Candidate generation}

The generator works in an environment that contains the target software, so it
can write and test code while constructing a candidate. Its prompt provides a
domain guide, a small set of expert-written seeds, and recent gate outcomes. It
must return the question, options, proposed answer, and runnable solution in a
fixed file format. The prompt also asks for distractors based on realistic
modeling or implementation errors. The saved solution is checked independently;
the generator's own execution is not treated as evidence that the candidate is
correct.

\subsection{Acceptance gates}

\paragraph{Local verification.}
A verifier runs the submitted solution in the target environment against an
input copy that omits the proposed answer. The candidate passes only if the
script finishes, prints exactly one answer, and that answer matches the
generator's proposal. This gate checks that the saved computation reproduces
the proposed answer; it does not judge whether the question is difficult.

\paragraph{No-tool screening.}
A no-tool screen is one model at one reasoning setting. It receives only the
question and options, with no files, code execution, retrieval, submitted
solution, or target software. The model answers three independently randomized
presentations of the candidate, and grading maps each selected value back to
the original option. If the model answers correctly in at least two calls, the
candidate is rejected as too easy under that screen. A candidate passes the
no-tool gate only if it passes every specified screen.

\paragraph{Tool-enabled solving.}
The remaining candidates go to an agent in a fresh workspace with the target
software and code execution. The agent sees the public question and options,
but not the proposed answer or the generator's solution. It must implement its
own solution and return one option within a fixed time limit. Candidates it
solves are accepted; the rest are flagged for review. This final gate separates
questions that are hard without the software from questions that are simply
underspecified, broken, or beyond the tested agent's budget.

\subsection{Feedback and recorded evidence}

Every attempt, including every rejection, is appended to the run database.
Before the next round, the generator receives recent outcomes and a summary of
repeated local-verification failures, and is asked to change approach when a
failure recurs. Only these in-loop outcomes enter the generator's prompt;
post-hoc audits do not affect generation.

Passing remains relative to the reported test conditions. A stronger no-tool
model may solve a retained candidate, while a different tool agent or time
budget may change the final result. ToolGate therefore saves the model,
reasoning setting, option presentation, time limit, and outcome for every call.
The retained item and this record are released together.

\section{FEniCSx Experimental Setup}
\label{sec:fenics}

\subsection{Tasks and seeds}

We study the FEniCSx ecosystem~\citep{baratta2023dolfinx}:
DOLFINx for finite-element solves, UFL for variational
forms~\citep{alnaes2014ufl}, and Basix for element definition and
tabulation~\citep{scroggs2022basix}. Its workflows combine mesh construction,
finite-element spaces, numerical assembly, and post-processing. They produce
precise answers that are difficult to obtain by hand but can be checked by
running modest programs.

The generator receives five expert-written seed items. All five are variants
of a nonlinear Poisson problem on a punctured square, with different forcing,
boundary conditions, and quantities of interest. The generator is not limited
to these templates. It produces tasks involving deformed meshes, variational
problems, quadrature, and Basix element operations.

\subsection{Models and execution settings}

Table~\ref{tab:protocol} lists the models used in the construction loop. The
generator, submitted solution scripts, and tool solver have access to the
FEniCSx environment. The no-tool screen uses direct API calls and receives only
the public question and options.

\begin{center}
\centering
\small
\begin{tabular}{@{}p{0.27\columnwidth}p{0.65\columnwidth}@{}}
\toprule
Role & Model and interface \\
\midrule
Generator & GPT-5.5, Codex CLI, medium reasoning \\
No-tool screen & GPT-5.4 API, reasoning disabled \\
Tool solver & GPT-5.5, Codex CLI, medium reasoning \\
\bottomrule
\end{tabular}
\captionof{table}{Model roles in the original FEniCSx construction loop. The generator
and tool solver can execute code and access FEniCSx; the no-tool screen cannot.}
\label{tab:protocol}
\end{center}

The no-tool screen makes three calls and rejects a candidate when at least two
are correct. The generator, each submitted solution, and each tool-solver run
have a 300-second wall-clock limit. The solver works in a fresh workspace and
returns its answer through a single-answer file. The generator receives outcome
counts, the 20 most recent database records, and a summary of repeated local
failures.

\subsection{End-to-end example}

The following is the full public question for one generated candidate
(item 0023):

\begin{quote}\small
In DOLFINx, create a $17{\times}14$ triangular unit-square mesh with right
diagonals. Before moving any coordinates, mark the original left boundary
$x=0$ and top boundary $y=1$ as Dirichlet facets, and the original right
boundary $x=1$ and bottom boundary $y=0$ as Neumann facets. Move each original
coordinate $(x,y)$ to $(X,Y)$ using
\[
X=x+0.041\sin(2\pi y)+0.017xy-0.009y^2+0.004x^2y,
\]
\[
\begin{array}{rl}
Y={}&y+0.027\sin(\pi x)\sin(2\pi y)+0.019x^2\\
&{}-0.014xy+0.006y^2.
\end{array}
\]
On the moved mesh, solve in the continuous degree-one Lagrange space for
$u_h$ satisfying
\[
\int a\nabla u_h\!\cdot\!\nabla v\,dx
=\int fv\,dx+\int g v\,ds
\]
on the marked Neumann facets, with $u_h=u_D$ on the marked Dirichlet facets.
Use the moved coordinates in all coefficient functions:
\[
\begin{array}{rl}
a={}&0.88+0.24X-0.13Y+0.05XY\\
&{}+0.031X^2+0.017Y^2,\\
f={}&1.07-0.36X+0.29Y+0.18XY\\
&{}-0.11X^2+0.07Y^2+0.05X^2Y,\\
g={}&-0.09+0.14X+0.06Y-0.035XY\\
&{}+0.022Y^2,\\
u_D={}&0.16+0.31X-0.27Y+0.05XY\\
&{}+0.028X^2-0.034Y^2.
\end{array}
\]
After solving, compute
\[
\begin{array}{rl}
J={}&\displaystyle\int_{\Omega_h}\big[(0.71-0.12X+0.18Y\\
&{}+0.064XY+0.027X^2-0.021Y^2)u_h^2\\
&{}+0.042|\nabla u_h|^2\\
&{}+0.018u_h(X-0.52Y+0.13XY)\big]\,dx.
\end{array}
\]
Use quadrature degree 5 for the volume and boundary forms. Report $J$ rounded
to exactly six decimal places. \\
(A)~0.058996 \quad (B)~0.060723 \quad (C)~0.060074 \quad (D)~0.061291
\end{quote}

The database records the complete in-loop path for this candidate:
\begin{itemize}
\item local verification runs the submitted script and returns (B), matching
  the proposed answer;
\item the original GPT-5.4 no-tool screen answers (C) in all three fixed-order
  calls, for zero correct answers; and
\item the independent GPT-5.5 Codex CLI solver uses FEniCSx and returns (B), so
  the candidate is accepted by the construction loop.
\end{itemize}

\section{Results}
\label{sec:experiments}

\subsection{Construction run}

We ran the generator for 500 rounds against a fresh FEniCSx database. Local
verification rejected 22 candidates whose scripts failed or did not reproduce
their proposed answers, leaving 478 (95.6\%). The original in-loop no-tool
screen used GPT-5.4 with reasoning disabled. It marked 260 of the 478
candidates as easy and rejected them. The remaining 218 entered the tool gate,
where the GPT-5.5 Codex CLI agent solved 212 with FEniCSx and failed on six.

\begin{figure}[t]
\centering
\includegraphics[width=\columnwidth]{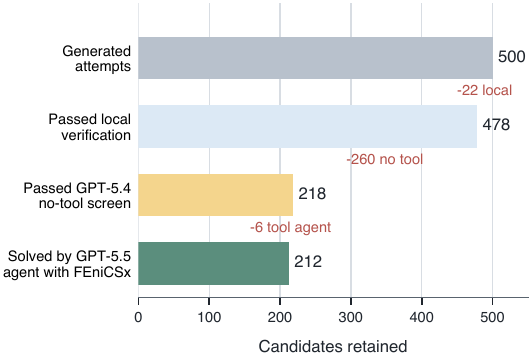}
\caption{The original construction run. Of 500 generated candidates, 478 pass
local verification, 218 pass the in-loop GPT-5.4 no-tool screen, and the
GPT-5.5 Codex CLI agent, configured for medium reasoning and given access to
FEniCSx, solves 212. These are the outcomes recorded during generation; later
audits define the stricter reported pool.}
\label{fig:construction}
\end{figure}

These counts describe the construction run exactly, including the feedback
seen by the generator. They do not define the final reported pool. The
original no-tool screen used a fixed answer order; a later audit found that
answer position affected its decisions. Section~\ref{sec:audits} reports that
audit, the corrected no-tool screens, and the resulting conservative pool.

\subsection{Resource use}

Five expert-written seeds and one domain guide are shared across the full run.
The 500 generator calls consume 29.4 agent-hours, and the 218 tool-solve calls
made during the construction loop consume 5.49 agent-hours. These are compute
measurements, not a controlled comparison with human authors. They show the
operational change introduced by ToolGate: experts define the domain and review
the output, while the pipeline performs repeated implementation, execution,
and screening.

\begin{table}[t]
\centering
\small
\begin{tabular}{@{}lr@{}}
\toprule
Stage & Measured resource \\
\midrule
Generator (500 calls) & 29.4 agent-h \\
In-loop tool solve (218 calls) & 5.49 agent-h \\
Each no-tool screen & 3 calls/item \\
\bottomrule
\end{tabular}
\caption{Measured resource use for the construction loop. Times are agent
wall-clock totals; local verification uses only local compute.}
\label{tab:resources}
\end{table}

\section{Protocol Audits and Corrections}
\label{sec:audits}

The original construction loop used a fixed answer order. Its GPT-5.4 screen
sent 218 candidates to the tool gate, which accepted 212. We report this as a
pilot outcome, not as the final yield, because the following audit shows that
the no-tool decision was partly driven by answer position. The corrected and
cross-family results, the stronger no-tool screen, and the final reported pool
were all computed after generation and never entered the generator's feedback
memory. Together, these post-generation checks leave 130
tool-solvable candidates and 128 unique protocol survivors after exact
deduplication. The following subsections explain how this conservative pool is
formed.

\subsection{Answer presentation}

The original fixed-order screen was sensitive to answer position. The
generator placed the proposed answer at C in 43.9\% of verified candidates,
while GPT-5.4 selected C in 66\% of its calls. Its per-call accuracy was 86.0\%
when C was correct and 29.4\% otherwise. We therefore repeated the screen with
independently shuffled options and value-based grading; positional accuracy
then ranged from 28\% to 33\%. Under this corrected protocol, 191 of the
pilot's 260 fixed-order ``easy'' labels no longer remain easy. All
post-generation results use the corrected protocol.

Randomization removes the letter cue but not every answer cue. The proposed
answer is one of the two middle numeric values in 80\% of verified candidates,
so a middle-value heuristic succeeds on 40.0\%. Future runs should balance
distractor ranks as well as randomize their displayed positions. We also use
fresh permutation seeds for release so that the retained pool is not tied to
the screening permutations used in this study.

\subsection{Cross-family validation}

\paragraph{No-tool screening.}
After correcting the GPT-5.4 presentation, we ask whether the result depends
on the OpenAI model family. We screen all 478 locally verified candidates with
the held-out Claude Opus 4.8 family. As with GPT-5.4, Opus receives three direct
API calls per candidate, with independently shuffled options and no files,
code execution, or FEniCSx. A model solves a candidate if at least two of its
three answers are correct. Of the 351 candidates that the shuffled GPT-5.4
screen fails to solve, Opus also fails on 256 (72.9\%). Of the 127 that the
shuffled GPT-5.4 screen solves, Opus also solves only 55 (43.3\%). Overall, the
models agree on 311 of 478 candidates (65.1\%) and disagree on 167 (34.9\%). We
use the 256 candidates that neither model solves as the input to the stronger
GPT-5.5 screen.

\paragraph{Tool-enabled solving.}
Tool-enabled solvability transfers more strongly across model families. We run
the held-out Claude Opus 4.8 tool agent on the same 212 candidates
already solved by the GPT-5.5 tool agent. Opus solves 209 (98.6\%); its three
failures are timeouts, not wrong answers. Thus, nearly every candidate solved
by the GPT-5.5 agent is also solved by the Opus agent. Table~\ref{tab:cross-family}
places the no-tool and tool-agent transfer rates side by side.

\begin{table}[t]
\centering
\small
\begin{tabular}{@{}p{0.47\columnwidth}p{0.45\columnwidth}@{}}
\toprule
OpenAI outcome & Matching Opus outcome \\
\midrule
GPT-5.4 shuffled no-tool fails (351) & Opus shuffled no-tool fails: 256 (72.9\%) \\
GPT-5.4 shuffled no-tool solves (127) & Opus shuffled no-tool solves: 55 (43.3\%) \\
GPT-5.5 tool agent solves (212) & Opus agent solves: 209 (98.6\%) \\
\bottomrule
\end{tabular}
\caption{Cross-family transfer under the no-tool and tool-enabled protocols.
The no-tool rows cover all 478 locally verified candidates. The tool-agent row
uses a cohort selected because GPT-5.5 solved it, so it measures one-way
transfer to Opus rather than an unbiased agent ranking. The original
fixed-order GPT-5.4 results are not used in this table.}
\label{tab:cross-family}
\end{table}

The Opus audit consumes 4.31 agent-hours, with a median of 61.8 seconds per
item. Together, the no-tool and tool-agent checks show that model family can
change which candidates appear difficult, while tool-enabled solvability
transfers almost completely on this cohort.

\subsection{Stronger no-tool reasoning}
\label{sec:compute}

Our goal is not merely to find candidates that defeat inexpensive screens. A
question is useful for evaluating scientific tool use only if a strong model
still cannot solve it from the prompt alone. The first two no-tool screens
identify questions answerable from prior knowledge, option cues, or limited
unaided reasoning, leaving 256 candidates for the stronger check. Passing those screens does
not yet show that software is needed: a stronger model may still derive the
answer with more reasoning but without FEniCSx.

This rescreen is a post-generation reporting filter, not part of the original
generation loop.
We therefore apply GPT-5.5 at medium reasoning---the API default when the
request omits the \texttt{reasoning\_effort} field---while continuing to
withhold files, code execution, and FEniCSx. It solves 121 of the 256 candidates by
majority. These candidates do not satisfy the stricter post-generation
no-tool criterion and are excluded from the reported pool. The remaining 135
pass this third no-tool screen.

The 135 candidates then enter the final tool check. Of these, 68 had already
passed the in-loop tool gate. We send the other 67 to the same GPT-5.5 Codex
CLI agent with FEniCSx; it solves 62 and fails on five. The reported pool
therefore contains 130 tool-solvable candidates. Exact question deduplication
removes two copies, leaving 128 unique protocol survivors.
Table~\ref{tab:postgen-audit} summarizes this corrected post-generation flow.

\begin{table}[t]
\centering
\small
\begin{tabular}{@{}lrrr@{}}
\toprule
Stage & In & Removed & Out \\
\midrule
Local verification (audit start) & 478 & --- & 478 \\
Shuffled GPT-5.4 no-tool & 478 & 127 solved & 351 \\
Opus 4.8 no-tool & 351 & 95 solved & 256 \\
GPT-5.5 medium no-tool & 256 & 121 solved & 135 \\
GPT-5.5 + FEniCSx & 135 & 5 failed & 130 \\
Exact deduplication & 130 & 2 duplicates & 128 \\
\bottomrule
\end{tabular}
\caption{Post-generation audit and final reported pool. The corrected audit
restarts from all 478 locally verified candidates because the original
fixed-order screen was position-biased. ``Solved'' means exclusion in the
no-tool rows; ``failed'' means exclusion in the tool-enabled row.}
\label{tab:postgen-audit}
\end{table}

As a diagnostic, we also run GPT-5.5 with reasoning explicitly disabled on the same 256
candidates; it solves 49 by majority. This run does not affect acceptance. The
two runs are stochastic and their solved sets are not nested. Neither run
affected the original construction loop or the generator's memory.

Among the 212 candidates accepted by the original construction loop, 115
remain unsolved by both randomized no-tool screens. On this fixed diagnostic
cohort, direct GPT-5.5 calls solve 16 with reasoning disabled and 47 at medium
reasoning (the API default), while the Codex CLI agent at medium reasoning with access to FEniCSx
solves all 115. These are parallel measurements on the same cohort, not a
sequential filter; the reasoning-disabled run does not affect acceptance.

\section{Related Work}
\label{sec:related}

We do not propose a new agent or a new scientific task. Our contribution is
upstream: an acceptance pipeline for generated evaluation candidates. It puts
three forms of evidence on each retained item---local verification, no-tool
failure, and tool-enabled success.

\paragraph{Expert authorship and automatic construction.}
GPQA~\citep{rein2024gpqa} and
ScienceAgentBench~\citep{chen2024scienceagentbench} are reference points for
expert-built scientific evaluation. ToolGate retains experts for domain design
and final review, but moves repeated execution and difficulty screening into a
shared automated process. HellaSwag~\citep{zellers2019hellaswag} established
adversarial filtering, AutoBencher~\citep{li2024autobencher} searches for items
that expose model failures, and AutoCodeBench~\citep{chou2025autocodebench}
checks generated programming tasks by sandboxed execution. ToolGate combines
these ideas around a scientific oracle: the software reproduces the label, a
no-tool model must fail, and a tool-enabled agent must succeed. Because static
benchmarks decay through contamination and saturation~\citep{white2024livebench},
the executable pipeline also supports repeated generation and re-screening.

\paragraph{Tool-use data and scientific agents.}
APIGen~\citep{liu2024apigen} is the closest generic precedent: it filters
generated function-calling data through format, execution, and LLM-judged
checks, while EigenData~\citep{chen2026eigendata} synthesizes and audits
function-calling environments. Both target general tool-use data rather than
scientific evaluation. Execution-based agent benchmarks such as
$\tau^2$-bench~\citep{barres2025tau2bench} verify environment state;
BrowseComp~\citep{wei2025browsecomp} exposes the value of an external tool, and
ToolFailBench~\citep{soni2026toolfailbench} constructs tool-required tasks and
controls. ToolGate instead makes no-tool failure and tool-enabled success
per-item acceptance criteria. Scientific-agent suites---SciCode,
CORE-Bench, and SciAgentArena~\citep{tian2024scicode,siegel2024corebench,liu2026sciagentarena}---evaluate
agents on curated or research-derived tasks, and ChemCrow~\citep{bran2023chemcrow}
demonstrates scientific work with expert tools. Our contribution is upstream:
generating and filtering new atomic candidates whose answers are reproduced by
the scientific software itself.

\section{Discussion}
\label{sec:discussion}

The study supports one central claim: an executable generation-and-acceptance
pipeline can construct candidates that exhibit a measured tool gap. Of the 478
locally verified candidates, the sequential no-tool screens exclude 343,
leaving 135; the specified tool-enabled agent solves 130 of them, and exact
deduplication leaves 128 unique protocol survivors. Generation and executable
label verification alone would therefore overstate the useful yield. The
acceptance layer is the main contribution: it moves repeated answer computation
and difficulty screening out of the per-item expert loop while retaining a
record of why each candidate passed. This conclusion, however, is only as
reliable as the gates used to measure it.

\paragraph{No-tool difficulty is protocol-relative.}
The no-tool gate, like any measurement instrument, is protocol-dependent and
susceptible to artifacts. Fixed option order let two letter priors---the generator's
and the screener's---masquerade as difficulty structure. Randomized,
value-graded screening removes the letter cue, but model family still changes
which candidates survive: GPT-5.4 and Opus disagree on 167 of 478 candidates
(34.9\%) under the randomized screens. By contrast, on the selected
tool-enabled cohort, 209 of the 212 GPT-5.5 successes also transfer to Opus.
No-tool failure is therefore evidence under named models and presentation
rules, not an intrinsic property of an item. Reporting cue-exploiting baselines
makes residual artifacts visible, and value-rank balance must be imposed at
generation time because no post-hoc permutation can repair it.

\paragraph{Gate ordering controls cost.}
The local-verification gate uses local compute, and each short no-tool screen
uses three direct calls with no workspace. The medium-reasoning screen uses a
larger inference budget, so in the corrected audit we run it only on the 256
candidates that survive the lighter screens, not all 478. This ordering is a
practical design principle for future production runs: expensive measurements
are reserved for fewer candidates. The supplementary material reports the
complete reasoning-mode protocols, outcome records, set overlaps, and token
accounting.

\paragraph{Outcome memory does not ensure diversity.}
Outcome memory tells the generator what recently failed, but it does not ask
for diversity. The earlier 212-candidate fixed-order in-loop cohort
concentrates in two
families: 147 Basix interpolation items and 63 DOLFINx deformed-mesh items,
together covering 210 of 212 candidates.
At token Jaccard similarity $\geq 0.7$, the largest cluster contains 66\% of
them, and five items are exact repeats. In the reported 130 protocol survivors,
exact question deduplication removes two copies and leaves 128 unique.
Seed examples are therefore steering controls rather than prerequisites. To
target other FEniCSx problem types or topics, a run can replace or rotate the
seed set; the pipeline can also start with no seed examples and let the
generator adapt its proposals from accumulated gate outcomes. Neither option
by itself guarantees coverage. Future production runs should combine such
seed policies with topic quotas and novelty penalties against accepted
history. Yield without concentration statistics is not enough.

\paragraph{Limitations.}
The acceptance result is protocol-relative, not absolute. A different model,
inference budget, or majority threshold can move candidates across the gate.
We also compare different scaffolds: the no-tool screen is a direct call,
whereas the tool-enabled solver is a multi-turn agent with a workspace. The
reported gap therefore bundles tool access with iteration and scaffolding; it
is not a causal estimate of tool access alone. We do not measure human
difficulty.

The protocol-surviving pool is selected for tool-enabled success, so it is not
designed to rank the same strong agents that define it. The multiple-choice format has a
25\% guessing floor and leaks magnitude information through numeric options.
Six-decimal answers also make some items computation-required because of
precision rather than conceptual depth. All candidates come from one
generator family and one domain, and the run concentrates on two templates.
Transfer to other tools remains future work.

The three gates establish operational rather than complete semantic validity.
They test whether the submitted script reproduces the proposed answer, whether
specified models fail without tools, and whether a specified tool-enabled agent
succeeds. A post-hoc audit identified a recurring FEniCSx data-layout defect in
97 of the 130 pre-deduplication protocol survivors and left five cases
unresolved; 28 were not implicated by this specific check, but were not thereby
proved correct. This finding does not change the recorded gate outcomes, but it
shows that executable label reproduction alone cannot detect every mismatch
between a question and its intended computation. Because the observed defect
is recurring and mechanically characterizable, it appears amenable to an
additional domain-specific executable gate. More generally, ToolGate is
modular: new semantic checks can be inserted as failure modes are identified.

\paragraph{Availability.}
We will release the candidates with solution scripts, gate records, and audit
flags; the run database including rejected candidates; screening logs; and
the pipeline code. These artifacts make the pipeline and its observed failure
modes reproducible and support the addition of further gates.

\section{Conclusion}

We presented ToolGate, an executable acceptance pipeline for generated
scientific evaluation candidates. A generator proposes each item; a solution
script must reproduce its proposed answer, randomized no-tool calls must fail,
and a tool-using agent must succeed. In 500 FEniCSx attempts, 478 candidates
reproduce their proposed answers. The sequential no-tool screens exclude 343,
and the full reported protocol leaves 130 candidates before exact deduplication
and 128 unique protocol survivors after it. This reduction is the main result:
generation is fast, but tool dependence must be measured and selected for.
ToolGate shifts repeated software execution and difficulty screening from
per-item expert labor into an auditable process. Each result remains tied to
named models, budgets, and presentation rules, so the same process can be rerun
as those conditions change.

\section*{Ethical Statement}

This work generates synthetic scientific evaluation questions, which can be
mistaken or misleading if released without their status records. We report all
gate outcomes and retain expert review as a release step. The items use standard open-source
scientific software and add no domain-specific dual-use capability.

\bibliography{aaai2027}

@inproceedings{rein2024gpqa,
  title={{GPQA}: A Graduate-Level {Google}-Proof {Q\&A} Benchmark},
  author={Rein, David and Hou, Betty Li and Stickland, Asa Cooper and Petty, Jackson and Pang, Richard Yuanzhe and Dirani, Julien and Michael, Julian and Bowman, Samuel R.},
  booktitle={First Conference on Language Modeling},
  year={2024},
}

@inproceedings{zellers2019hellaswag,
  title={{HellaSwag}: Can a Machine Really Finish Your Sentence?},
  author={Zellers, Rowan and Holtzman, Ari and Bisk, Yonatan and Farhadi, Ali and Choi, Yejin},
  booktitle={Proceedings of the 57th Annual Meeting of the Association for Computational Linguistics},
  pages={4791--4800},
  year={2019},
}

@inproceedings{li2024autobencher,
  title={{AutoBencher}: Towards Declarative Benchmark Construction},
  author={Li, Xiang Lisa and Kaiyom, Farzaan and Liu, Evan Zheran and Mai, Yifan and Liang, Percy and Hashimoto, Tatsunori},
  booktitle={International Conference on Learning Representations},
  year={2025},
}

@article{chou2025autocodebench,
  title={{AutoCodeBench}: Large Language Models are Automatic Code Benchmark Generators},
  author={Chou, Jason and Liu, Ao and Deng, Yuchi and Zeng, Zhiying and Zhang, Tao and others},
  journal={arXiv preprint arXiv:2508.09101},
  year={2025},
}

@article{liu2024apigen,
  title={{APIGen}: Automated Pipeline for Generating Verifiable and Diverse Function-Calling Datasets},
  author={Liu, Zuxin and Hoang, Thai and Zhang, Jianguo and Zhu, Ming and Lan, Tian and Kokane, Shirley and Tan, Juntao and Yao, Weiran and Liu, Zhiwei and Feng, Yihao and others},
  journal={arXiv preprint arXiv:2406.18518},
  year={2024},
}

@inproceedings{chen2024scienceagentbench,
  title={{ScienceAgentBench}: Toward Rigorous Assessment of Language Agents for Data-Driven Scientific Discovery},
  author={Chen, Ziru and Chen, Shijie and Ning, Yuting and Zhang, Qianheng and Wang, Boshi and Yu, Botao and Li, Yifei and Liao, Zeyi and Wei, Chen and Lu, Zitong and others},
  booktitle={International Conference on Learning Representations},
  year={2025},
}

@article{tian2024scicode,
  title={{SciCode}: A Research Coding Benchmark Curated by Scientists},
  author={Tian, Minyang and Gao, Luyu and Zhang, Shizhuo Dylan and others},
  journal={arXiv preprint arXiv:2407.13168},
  year={2024},
}

@article{siegel2024corebench,
  title={{CORE-Bench}: Fostering the Credibility of Published Research Through a Computational Reproducibility Agent Benchmark},
  author={Siegel, Zachary S. and Kapoor, Sayash and Nadgir, Nitya and Stroebl, Benedikt and Narayanan, Arvind},
  journal={arXiv preprint arXiv:2409.11363},
  year={2024},
}

@article{white2024livebench,
  title={{LiveBench}: A Challenging, Contamination-Limited {LLM} Benchmark},
  author={White, Colin and Dooley, Samuel and Roberts, Manley and Pal, Arka and Feuer, Ben and Jain, Siddhartha and Shwartz-Ziv, Ravid and Jain, Neel and Saifullah, Khalid and Dey, Sreemanti and others},
  journal={arXiv preprint arXiv:2406.19314},
  year={2024},
}

@article{wei2025browsecomp,
  title={{BrowseComp}: A Simple Yet Challenging Benchmark for Browsing Agents},
  author={Wei, Jason and Sun, Zhiqing and Papay, Spencer and McKinney, Scott and Han, Jeffrey and Fulford, Isa and Chung, Hyung Won and Passos, Alex Tachard and Fedus, William and Glaese, Amelia},
  journal={arXiv preprint arXiv:2504.12516},
  year={2025},
}

@article{chen2026eigendata,
  title={{EigenData}: A Self-Evolving Multi-Agent Platform for Function-Calling Data Synthesis, Auditing, and Repair},
  author={Chen, Jiaao and Qi, Jingyuan and Gao, Mingye and Wang, Wei-Chen and Wang, Hanrui and Jin, Di},
  journal={arXiv preprint arXiv:2603.05553},
  year={2026},
}

@article{barres2025tau2bench,
  title={$\tau^2$-Bench: Evaluating Conversational Agents in a Dual-Control Environment},
  author={Barres, Victor and Dong, Honghua and Ray, Soham and Si, Xujie and Narasimhan, Karthik},
  journal={arXiv preprint arXiv:2506.07982},
  year={2025},
}

@article{soni2026toolfailbench,
  title={{ToolFailBench}: Diagnosing Tool-Use Failures in {LLM} Agents},
  author={Soni, Harsh},
  journal={arXiv preprint arXiv:2607.04686},
  year={2026},
}

@article{liu2026sciagentarena,
  title={Benchmarking {AI} Agents for Addressing Scientific Challenges Across Scales},
  author={Liu, Tianyu and Wang, Allen Xin and Panescu, Antonia and Chen, Lisa Xinyi and Long, Wenxin and Wei, Xinyu and Jing, Yueqian and Zeng, Ziyao and Chen, Jihang and Jiang, Sihan and others},
  journal={arXiv preprint arXiv:2606.12736},
  year={2026},
}

@article{baratta2023dolfinx,
  title={{DOLFINx}: The Next Generation {FEniCS} Problem Solving Environment},
  author={Baratta, Igor A. and Dean, Joseph P. and Dokken, J{\o}rgen S. and Habera, Michal and Hale, Jack S. and Richardson, Chris N. and Rognes, Marie E. and Scroggs, Matthew W. and Sime, Nathan and Wells, Garth N.},
  journal={Journal of Open Source Software},
  volume={8},
  number={84},
  pages={5120},
  year={2023},
}

@article{alnaes2014ufl,
  title={Unified Form Language: A Domain-Specific Language for Weak Formulations of Partial Differential Equations},
  author={Aln{\ae}s, Martin S. and Logg, Anders and {\O}lgaard, Kristian B. and Rognes, Marie E. and Wells, Garth N.},
  journal={ACM Transactions on Mathematical Software},
  volume={40},
  number={2},
  pages={1--37},
  year={2014},
}

@article{bran2023chemcrow,
  title={Augmenting Large Language Models with Chemistry Tools},
  author={Bran, Andres M. and Cox, Sam and Schilter, Oliver and Baldassari, Carlo and White, Andrew D. and Schwaller, Philippe},
  journal={Nature Machine Intelligence},
  volume={6},
  pages={525--535},
  year={2024}
}

@article{scroggs2022basix,
  title={Basix: a runtime finite element basis evaluation library},
  author={Scroggs, Matthew W. and Baratta, Igor A. and Richardson, Chris N. and Wells, Garth N.},
  journal={Journal of Open Source Software},
  volume={7},
  number={73},
  pages={3982},
  year={2022}
}

\end{document}